\documentclass[sigconf,nonacm]{acmart}

\usepackage{booktabs}
\usepackage{multirow}
\usepackage{graphicx}
\usepackage{algorithm}
\usepackage{algpseudocode}
\usepackage{amsmath}
\usepackage{subcaption}
\usepackage{enumitem}

\AtBeginDocument{%
  \providecommand\BibTeX{{%
    \normalfont B\kern-0.5em{\scshape i\kern-0.25em b}\kern-0.8em\TeX}}}

\begin{document}

\title{
Embodied Human–Robot Interaction via Acoustics: A MARL Approach with AcoustoBots for Spatial Data Physicalization
}

\author{Shiqi Liu}
\affiliation{
\department{Department of Computer Science}
\institution{University College London}
\city{London}
\country{United Kingdom}}
\email{zczqs66@ucl.ac.uk}

\author{Narsimlu Kemsaram}
\affiliation{
\department{Department of Artificial Intelligence}
\institution{University of Malaya}
\city{Kuala Lumpur}
\country{Malaysia}}
\email{narsimlu.kemsaram@um.edu.my}

\author{Prateek Mittal}
\affiliation{
\department{Department of Computer Science}
\institution{University College London}
\city{London}
\country{United Kingdom}}
\email{prateek.mittal@ucl.ac.uk}

\author{Pengyuan Wei}
\affiliation{
\department{Department of Computer Science}
\institution{University College London}
\city{London}
\country{United Kingdom}}
\email{pengyuan.wei.22@ucl.ac.uk}

\author{Sriram Subramanian}
\affiliation{
\department{Department of Computer Science}
\institution{University College London}
\city{London}
\country{United Kingdom}}
\email{s.subramanian@ucl.ac.uk}



\begin{abstract}

Traditional data physicalization is often static and disconnected from real environments, limiting its ability to convey embodied spatial dynamics and engage users. To address this limitation, we present \textit{AcoustoBots}, a mobile acoustophoretic data-physicalization platform in which TurtleBot3 robots carry upward-facing $8\times8$ ultrasonic phased arrays. Each array levitates a particle whose height ($1$--$10$ cm) encodes a local urban scalar value, such as population density, noise, or traffic. A MARL (Multi-Agent Reinforcement Learning) policy based on the Multi-Agent Deep Deterministic Policy Gradient (MADDPG) algorithm, with centralized training and decentralized execution, selects collision-aware navigation actions, while a high-rate Gerchberg-Saxton-Phased Array of Transducers (GS-PAT) acoustic controller maintains trap stability and updates array phases to achieve the commanded height during motion. This creates a closed perception–display–action loop. We evaluate single-robot city-to-city traversal and dual-robot cooperative coverage on a 4 m x 3 m scaled UK map using PhaseSpace-based localization for repeatable multi-robot trials. Results show stable in-motion levitation and consistent, location-dependent height rendering, with task success rates of $90\%$ and $80\%$ for the single- and dual-robot regimes, respectively, over 10 trials per regime, and low collision counts. These findings support acoustophoretic levitation as a simple, glanceable, robot-mediated communication cue for embodied human–robot interaction in spatial analytics.

\end{abstract}



\begin{teaserfigure}
  \includegraphics[width=0.57\textwidth]{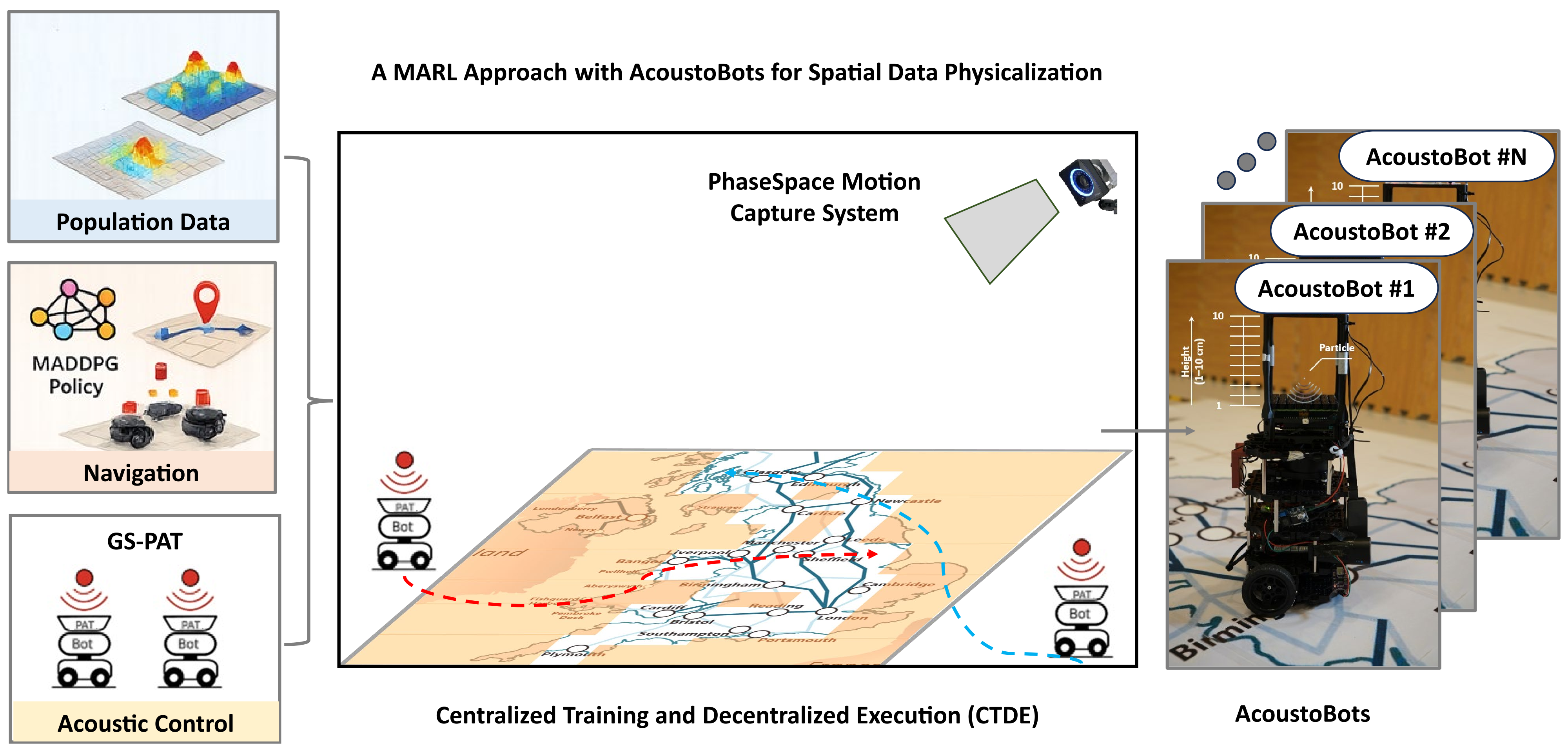}\centering
  \caption{
    Overview of the proposed AcoustoBots platform for embodied spatial data physicalization and robot-mediated communication. AcoustoBots navigate on a projected UK grid map and sample local urban scalar data, such as population density, noise, or traffic. Each TurtleBot3 carries an upward-facing 8×8 ultrasonic phased array that levitates a particle whose height (1–10 cm) encodes the local scalar value as a glanceable physical cue. A MARL policy selects safe and informative navigation actions from robot state and physicalization-related observations, while a high-rate GS-PAT-based acoustic controller continuously updates array phases to maintain stable levitation and control the commanded particle height during motion, thereby closing the perception–display–action loop.
  }
  \Description{
  }
  \label{fig:BlockDiagram}
\end{teaserfigure}

\maketitle

\section{INTRODUCTION}

\label{sec:introduction}

Cities increasingly depend on timely, spatially resolved information—such as noise exposure, traffic flow, air quality, and population dynamics—to guide planning and operations~\cite{batty2013new}. Yet the dominant paradigm still routes sensing through offline pipelines and then visualizes results on flat screens that are spatially detached from the environments they describe~\cite{zheng2014urban}. This separation places a burden on users to mentally re-project abstract maps and color gradients back onto the physical world, limiting embodied understanding and rapid sense-making in situ.

Data physicalization encodes information directly into perceivable physical form. 
Prior studies suggest that physical representations can outperform screen-based displays in information-retrieval tasks due to their tangibility, depth cues, and body-centric spatial reasoning~\cite{jansen2013evaluating}. Willett et al. distinguish situated representations—data placed near its referent—from embedded ones that spatially coincide with it, arguing that embedded representations support the most natural perception~\cite{willett2017embedded}. 
However, many physicalization systems remain static or location-bound acoustophoretic platforms~\cite{marzo2017ultraino}, while mobile robotic physicalizations (e.g., ShapeBots \cite{suzuki2019shapebots} and wheeled micro-robot composites \cite{le2018dynamic}) achieve mobility via surface contact without acoustic manipulation. As a result, today’s systems rarely achieve simultaneous (i) mobility, (ii) physicalization, and (iii) data-responsiveness that follows the user’s or robot’s focus of attention. Figure~\ref{fig:BlockDiagram} illustrates the proposed perception--display--action loop.


The next frontier for human–robot symbiosis and robot-mediated communication needs a platform that travels to the referent and renders data as an embodied signal in situ. We introduce \textit{AcoustoBots}, mobile robots that combine a TurtleBot3 base with an upward-facing $8\times8$ ultrasonic phased-array. Each robot moves across a scaled map and encodes local scalar field values (for example, population density, noise, or traffic) as the height of a levitated particle (1--10 cm). This creates a glanceable, co-located data cue that moves with the agent. A MARL policy converts local field context into navigation decisions. A real-time GS-PAT-based acoustic controller maintains trap stability and updates array phases to realize the commanded height during motion. 
This closes a perception--display--action loop in which coordination is learned via MADDPG algorithm~\cite{lowe2017multi} under centralized training and decentralized execution~\cite{amato2024introduction}, allowing robots to act while communicating local context through an interpretable height cue.
Achieving this coupling is technically nontrivial. 
First, acoustic traps are sensitive to robot acceleration, turning, and vibration, so stability must be maintained under continuous locomotion. 
Second, multi-robot coordination must balance exploration efficiency (coverage and time) with physicalization fidelity (correct height rendering at the correct location), while avoiding collisions and interference. 
Third, bridging the above first and second steps is challenging because the system combines two tightly coupled control layers—mobile navigation and acoustic manipulation—each with distinct update rates and uncertainties.
Finally, for human interpretation, the data channel must remain legible and co-located with the referent as robots move between city waypoints.

Our main contributions are: i) a mobile acoustophoretic physicalization platform, \emph{AcoustoBots}, that couples autonomous navigation with in-situ acoustic levitation, extending acoustophoretic physicalization from static installations to mobile, exploratory settings, ii) a real-time acoustic control module that maps scalar field values to discrete levitation heights and maintains trap stability via GS-PAT phase updates throughout motion, iii) a MARL coordination framework using the MADDPG algorithm with physics-informed reward design, enabling two AcoustoBots to cooperatively cover geographic waypoints while continuously rendering population data, and iv) a ROS~2-integrated experimental evaluation on a $4\,\mathrm{m}\times 3\,\mathrm{m}$ scaled UK map with PhaseSpace-based localization, showing robust navigation and embodied, location-dependent particle-height rendering, demonstrating two-robot coordination and initial multi-robot deployment.

This paper is organized as follows. Section~\ref{RelatedWork} reviews prior work on data physicalization, acoustic levitation, and multi-agent reinforcement learning. Section~\ref{ProblemStatement} formalizes the problem and tasks. Sections~\ref{Methodology} and~\ref{MADDPGModel} describe the proposed AcoustoBots system and the MADDPG-based training pipeline. Section~\ref{Evaluation} presents the experimental setup, quantitative results, and discussion of robustness, limitations, and future directions. Finally, Section~\ref{Conclusions} concludes the paper.

\section{RELATED WORK} \label{RelatedWork}

Our work lies at the intersection of data physicalization, embodied interaction, acoustophoretic levitation, multi-robot coordination, and multi-agent reinforcement learning. 

\subsection{Data Physicalization Representations}

Data physicalization encodes information through physical form, enabling interaction via depth cues and body-centric spatial reasoning~\cite{dragicevic2021data}. Empirical studies report advantages over screen-based visualization for information retrieval and spatial tasks~\cite{herman2025touching}. Prior work distinguishes \textit{situated} representations (near a referent) from \textit{embedded} representations (co-located with the referent), arguing that embedded physicalizations can provide more natural perception~\cite{bae2022making}. 
However, situated visualization research has largely focused on screen-based or AR approaches, while non-screen physical encodings remain comparatively underexplored~\cite{bressa2022situated}. Motivated by this gap, we explore a mobile, contactless physicalization approach: AcoustoBots navigate to city waypoints and render population values through levitated particles whose heights encode the local scalar field, creating an embedded representation that supports in-situ comprehension.

\subsection{Data Physicalization via Robotic Systems}

Robotic platforms enable reconfigurable physicalizations, for example through swarms that form shapes or charts and support interactive exploration~\cite{suzuki2019shapebots}, ~\cite{le2018dynamic}. 
Complementary to contact-based robotic physicalizations, acoustic levitation systems provide mid-air representations without surface contact~\cite{al2022non}, ~\cite{kemsaram2025acoustobots}. 
Our work combines these directions by adding \emph{mobility} and \emph{learning-based coordination}, using single-particle height as a simple and robust physical cue during navigation.

\subsection{Acoustics and Multi-Robot Coordination}

Acoustic levitation enables non-contact manipulation of microparticles~\cite{andrade2020acoustic}, \cite{becsevli2025sonarios} but accurate control remains challenging due to nonlinear field dynamics and sensitivity to disturbances~\cite{ochiai2014three}, \cite{vandaele2005non}. To extend workspace, prior work explores multi-emitter or multi-robot configurations where repositioning acoustic sources enables more flexible manipulation~\cite{paneva2023optimal}, \cite{kemsaram2025cooperative}. 
In contrast, we focus on stable in-motion levitation as an interpretable, human-facing cue: reliable height rendering while robots travel between city waypoints.


\subsection{MARL for Data Physicalization}

MARL has shown strong potential in cooperative control tasks such as multi-robot navigation~\cite{oroojlooy2023review}. MADDPG algorithm supports centralized training and decentralized execution, balancing global coordination during learning with local autonomy at runtime~\cite{lowe2017multi}. In our setting, MARL is suitable because policies must jointly optimize collision avoidance, route efficiency, and physicalization fidelity. The levitated height cue becomes part of the interaction loop rather than post-processing~\cite{szafir2021connecting}.


\section{PROBLEM STATEMENT} 
\label{ProblemStatement}


This work addresses a coupled challenge: enabling geographically contextualized navigation for multiple mobile acoustic robots while also providing an embodied data physicalization of urban population metrics. 

\begin{figure}[!htbp]
    \centering \includegraphics[width=0.45\textwidth]{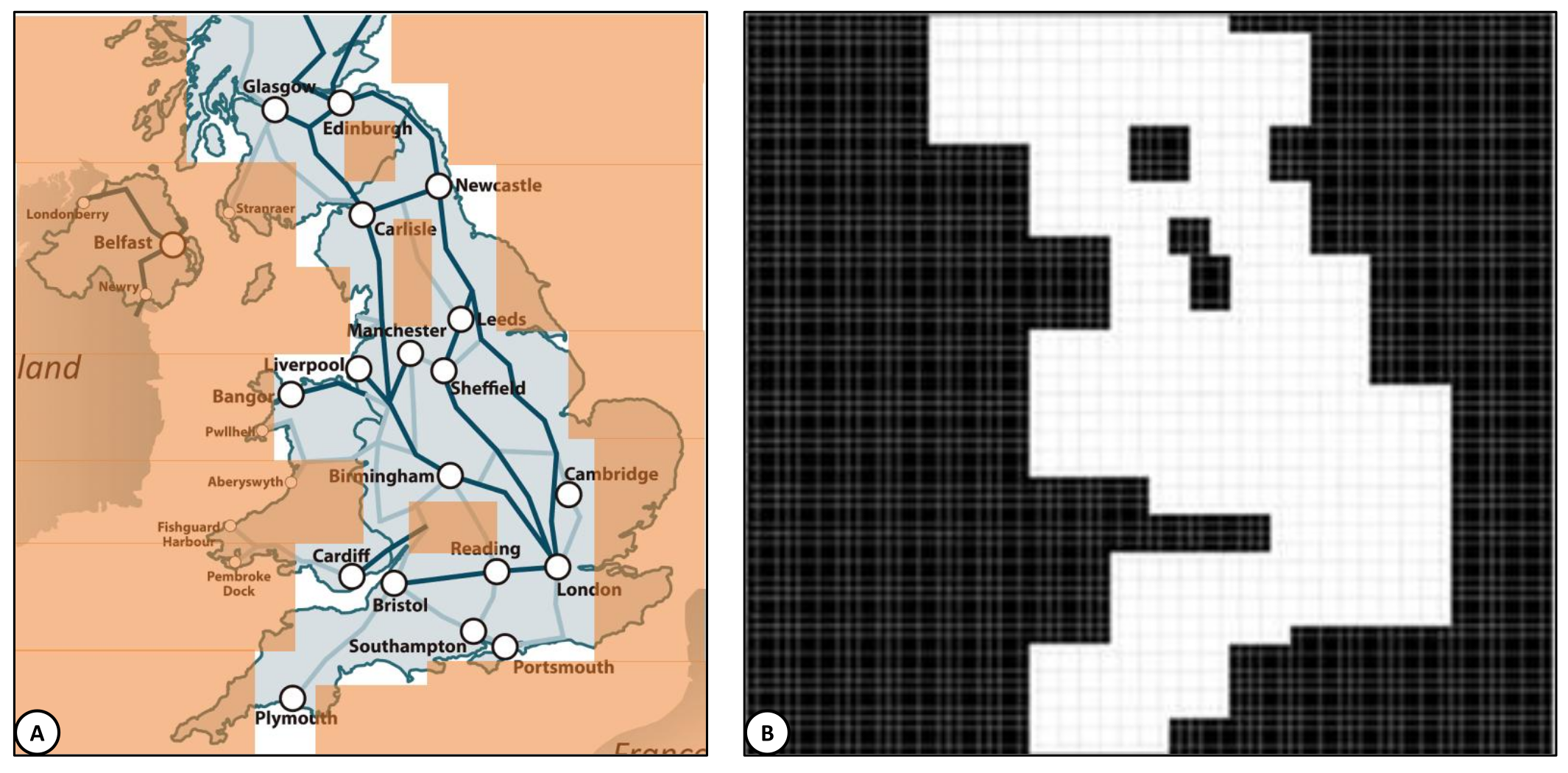}
    \caption{
    Navigation environment used for AcoustoBot training and evaluation: A) Scaled UK map with labeled city waypoints and inter-city routes, representing the physical scenario for city-to-city traversal and spatial data physicalization, and B) Corresponding grid-world abstraction used for MARL training, showing the traversable workspace and obstacle layout for learning collision-aware navigation toward target locations.
    }
    \label{fig:ProblemStatement}
    \Description{Navigation environment and grid abstraction used for training and evaluation.}
\end{figure}

\textit{Data physicalization and use case}: A scaled map of the UK is spread out on the ground, where two AcoustoBots are trained with MARL based on the MADDPG algorithm to navigate from one city to another (see Figure \ref{fig:ProblemStatement}). This setup requires decentralized coordination, collision avoidance, and integration of geographic constraints for effective multi-agent navigation. Each robot carries a phased array of transducers (PAT) board for acoustic levitation, allowing the system to encode demographic information in an interpretable way. A single particle is levitated above the robot, and its vertical displacement represents a scalar quantity such as city population, turning height into an intuitive physical metric that can be perceived in situ.


\textit{Workspace and field}: The projected UK map is modeled as a metric grid workspace $\mathcal{M}\subset\mathbb{R}^2$ with cell size $\Delta$. A scalar field $\phi:\mathcal{M}\to\mathbb{R}_{\ge 0}$ encodes population-related intensity over the map. A set of city centroids $\{c_k\}$ defines waypoints and evaluation locations for city-to-city navigation and coverage.



\textit{Embodied physicalization}: Each AcoustoBot comprises a TurtleBot3 base and an upward-facing $8{\times}8$ PAT board that levitates a single particle. At robot pose $\mathbf{x}_i=[x_i,y_i,\theta_i]^\top$, the commanded particle height encodes the local field value on a \emph{1--10\,cm} scale. A high-rate acoustic controller updates array phases to stabilize the trap and regulate particle height during motion, so the physical cue remains legible as the robot accelerates, turns, and stops.




\textit{Agents and dynamics}: We consider two agents $i\in\{1,2\}$ with unicycle kinematics:
\begin{equation}
\dot{x}_i=v_i\cos\theta_i,\quad \dot{y}_i=v_i\sin\theta_i,\quad \dot{\theta}_i=\omega_i
\label{eq:uni}
\end{equation}
subject to bounds $|v_i|\le v_{\max}$ and $|\omega_i|\le \omega_{\max}$. 


\textit{Tasks}: We study two representative tasks:
(1) \emph{Single-robot city-to-city traversal} with continuous height rendering of $\phi$ along the route, and (2) \emph{Two-robot informative coverage} in which the team coordinates to cover the map efficiently while emphasizing high-gradient regions of the scalar field (i.e., locations where population changes rapidly across neighboring cells), maintaining both navigation safety and physicalization fidelity.

Using this combination of projection-based context and levitated data markers, our approach grounds reinforcement learning–driven navigation in a tangible and interpretable physical display of the population distribution.

\section{PROPOSED SYSTEM} \label{Methodology}

\subsection{System Overview}



The proposed system establishes a closed loop of \emph{pose on the grid} $\rightarrow$ \emph{data sensing} $\rightarrow$ \emph{levitation-height physicalization} $\rightarrow$ \emph{policy decision} $\rightarrow$ \emph{motion with height regulation}. AcoustoBots equipped with a PAT board form the core of the proposed data physicalization interface.

\begin{figure}[!htbp]
    \centering \includegraphics[width=0.45\textwidth]{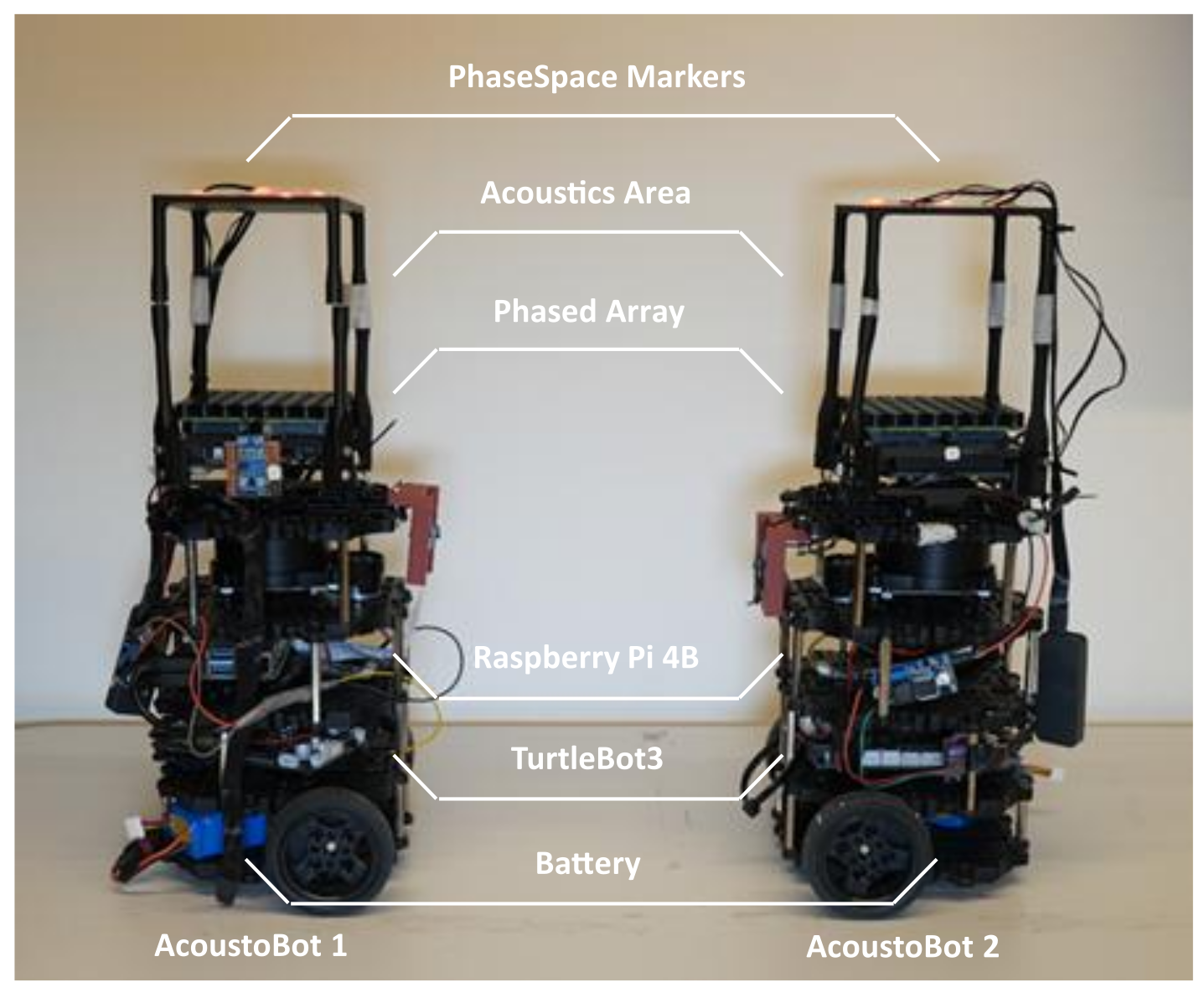}
    \caption{
    Hardware platform of the proposed TurtleBot3-based AcoustoBot. Each robot integrates onboard computing (Raspberry Pi 4B), a motor controller for mobile navigation, PhaseSpace markers for external localization, a battery power supply, and an upward-facing 8×8 ultrasonic phased array for acoustic levitation and spatial data physicalization. Two AcoustoBots are shown, illustrating the shared platform configuration used in the multi-robot experiments.
    }
    \label{SystemOverview}
\end{figure}

Each AcoustoBot is a TurtleBot3\footnote{\url{https://emanual.robotis.com/docs/en/platform/turtlebot3/overview/}} differential-drive base equipped with an upward-facing $8\times8$ phased-array transducer board (as shown in Figure \ref{SystemOverview}). As the robot moves across the map, local scalar values (e.g., population, noise, traffic) are encoded as a levitated particle height, creating a glanceable and co-located physical cue that supports robot-mediated communication in situ.


\subsection{System Architecture}

\begin{figure}[!htbp]
  \centering
  \includegraphics[width=0.45\textwidth]{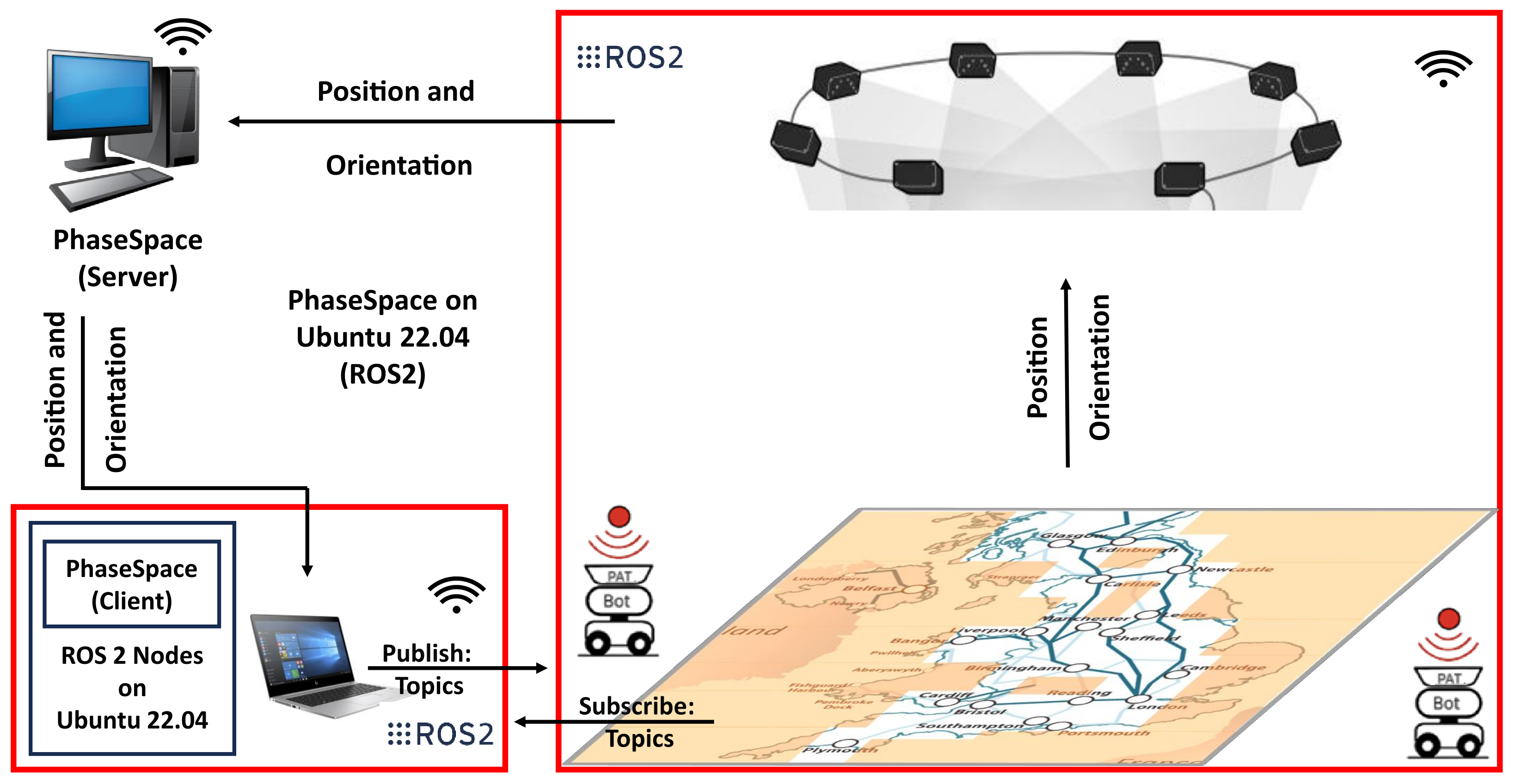}
  \caption{
  System architecture and ROS 2 data flow for the proposed AcoustoBots platform. A PhaseSpace motion-capture server estimates each AcoustoBot’s 6-DoF pose and streams the data to an Ubuntu client. A ROS 2 bridge node publishes synchronized pose topics to the network, enabling each AcoustoBot to localize itself, execute the learned MARL policy, and coordinate collision-aware navigation. Velocity commands are then sent to the robots for motion execution on the projected UK map.
  } \label{fig:SystemArchitecture}
  \Description{System architecture and ROS 2 data flow for the proposed AcoustoBots platform.}
\end{figure}

Figure~\ref{fig:SystemArchitecture} summarizes the ROS 2 localization, communication, and motion-command pipeline:



\begin{itemize}
  \item \texttt{mapper}: maintains an occupancy grid for avoidance, a scalar field layer $\phi$ (sampled/interpolated), and a coverage map $\Psi$ (novelty).
  \item \texttt{acoustic\_controller}: maps $\phi(\mathbf{x})$ to a height command $h^\star$, computes holographic phases for a single-focus trap at $z_f(h^\star)$, and regulates measured height $\hat h \rightarrow h^\star$ at high rate.
  \item \texttt{policy\_node}: executes the onboard actor $\pi_\theta(a|o)$ (centralized training and decentralized execution at runtime), consuming robot state, map context, and physicalized signals (e.g., height differentials). This is described in more detail in Section~\ref{MADDPGModel}.
  \item \texttt{base\_ctrl}: tracks motion commands $[v,\omega]$ with jerk limits.
  \item \texttt{safety\_supervisor}: enforces acoustic pressure caps and publishes health flags. Violations are fed back to the policy as penalties and can gate execution.
  \item \texttt{telemetry\_logger}: records pose, height, pressure, rewards, and events for evaluation.
\end{itemize}
All messages are time-stamped, and a shared clock ensures coherent fusion across sensing, perception, control, navigation, and logging.

This architecture supports real-time pose estimation, decentralized control, and collision-aware navigation. The PhaseSpace\footnote{\url{https://www.phasespace.com/software/}} motion capture system serves as a high-precision localization source (in place of IMU or GPS) in the robotics lab, where high-speed cameras track active LED markers and microdrivers attached to each AcoustoBot at 240\,Hz. Compared to conventional odometry-only localization, this provides millimeter-level positioning accuracy without cumulative drift, enabling repeatable evaluation of learned multi-robot behaviors.
A dedicated Ubuntu 22.04 client receives rigid-body pose data from the PhaseSpace server and persists it as JSON files in a shared memory space. A ROS~2 bridge node periodically reads these JSON files and publishes synchronized pose data on \texttt{/tb3\_1/pose} and \texttt{/tb3\_2/pose}. 

\subsection{System Design}

\subsubsection{Hardware design}

Each AcoustoBot consists of a TurtleBot3 differential-drive base equipped with i) an upward-facing $8\times8$ ultrasonic phased-array transducer board operating at 40\,kHz, ii) wheel encoders for local state estimation, and iii) onboard computing (Raspberry Pi 4B) running ROS~2 Humble on Ubuntu 22.04. 

\subsubsection{Software design}

The software design is structured into four layers, also known as the software stack: 

i) Data physicalization and acoustic control: 
Scalar values are physicalized as discrete height levels (1--10\,cm). Given a target height, a holographic GS-PAT solver\footnote{\url{https://github.com/RMResearch/OpenMPD\_Solvers/tree/main/GSPAT\_Solver}} computes phase configurations for each transducer to create a single-focus acoustic trap at the desired height. This enables non-contact manipulation where particle height is controlled via phase patterns rather than direct mechanical actuation. The \texttt{acoustic\_controller} updates phases at a high rate and maintains stable levitation during robot motion, while exposing pressure signals to the supervisor and logger.

ii) Perception and mapping: 
The \texttt{mapper} maintains a) an occupancy grid for collision avoidance, b) a scalar field layer $\phi$ (sampled and spatially interpolated), and c) a coverage map $\Psi$. Pose is estimated from odometry and PhaseSpace. All topics are synchronized in the ROS 2 framework to ensure consistent timing across nodes.

iii) Policy layer:
Onboard actors $\pi_\theta(a|o)$ run at 10--20\,Hz to generate motion commands for user-defined tasks (e.g., city-to-city traversal). The learned policy uses the robot state and physicalized signals (e.g., height differentials) to favor safe, efficient, and informative trajectories while remaining compatible with levitation constraints. 

iv) Safety supervisor:
A runtime guardian enforces acoustic pressure caps and publishes health flags used by \texttt{policy\_node}. The supervisor can gate execution when limits are approached and logs safety-related events for post-analysis.

\section{Policy Training} \label{MADDPGModel}


In this work, we utilize the \textit{MADDPG} algorithm \cite{lowe2017multi}, implemented in PyTorch using the PettingZoo parallel environment. MADDPG is adopted to support \emph{decentralized execution} on the robots while enabling \emph{coordinated behavior} during training, which is important for human-robot interaction, where robot actions must remain safe, interpretable, and consistent in shared spaces.

\subsection{Centralized Training with Decentralized Execution}

MADDPG extends deterministic policy gradients to MARL through Centralized Training with Decentralized Execution (CTDE). Each agent maintains its own actor--critic pair, while coordination is learned using a critic that conditions on the joint state/action during training. Execution remains decentralized: at runtime, each robot requires only its local observation and its learned actor.

\subsubsection{MADDPG formulation (CTDE)}

For each agent $i$, MADDPG maintains an actor $\mu_i(o_i|\theta_i)$ that maps local observations to actions, and a centralized critic $Q_i(x,a_1,\ldots,a_N|\phi_i)$ that conditions on joint information during training. We adopt CTDE to learn coordinated and collision-aware behavior while keeping execution decentralized on the robots.

Each agent stores transitions $(o_i,a_i,r_i,o'_i,d_i)$ in a replay buffer, and mini-batches are sampled synchronously across agents to update the centralized critics. During training, exploration noise is added to encourage diverse behavior, while at runtime each robot executes only its learned actor $\mu_i$ using local observations. Critic and actor updates follow the standard MADDPG losses, and target networks are updated via soft updates for stability.

\subsubsection{Algorithmic procedure}

The overall training procedure of MADDPG, along with its implementation\footnote{\url{https://github.com/Sakitama0227/MADDPG-clean}}, is summarized in Algorithm~\ref{alg:maddpg}. 




\begin{algorithm}[!htbp]
\caption{MADDPG Training Procedure}
\label{alg:maddpg}
\begin{algorithmic}[1]
\State Initialize actor $\mu_i$, critic $Q_i$, and target networks $\mu'_i, Q'_i$ for each agent $i$
\State Initialize replay buffer $B_i$ for each agent
\For{episode $=1$ to $M$}
    \State Reset environment and obtain initial observations $\{o_1,\dots,o_N\}$
    \For{$t=1$ to $T$}
        \For{each agent $i$}
            \State Select action $a_i \gets \mu_i(o_i) + \text{exploration noise}$
        \EndFor
        \State Execute joint action $\{a_1,\dots,a_N\}$, observe $r_i, o'_i, d_i$
        \State Store $(o_i, a_i, r_i, o'_i, d_i)$ into buffer $B_i$
        \If{replay buffer is sufficiently full}
            \For{each agent $i$}
                \State Sample mini-batch from all agents' buffers
                \State Compute target actions $a'_j \gets \mu'_j(o'_j)$ for all $j$
                \State Update critic by minimizing TD loss
                \State Update actor via policy gradient
            \EndFor
            \State Soft update target networks with parameter $\tau$
        \EndIf
    \EndFor
\EndFor
\end{algorithmic}
\end{algorithm}

\subsection{Environment Setup}

We design a multi-agent grid-based navigation environment using the PettingZoo parallel API.

\subsubsection{Environment design}

The environment is a 2D grid world of size $40\times40$ with static obstacles loaded from an external map (occupied cells ``\#''). Two agents $\{\text{agent}_1,\text{agent}_2\}$ start from predefined candidate positions and share a goal location $g$. Each agent has a discrete action space of size $|\mathcal{A}|=9$ consisting of four cardinal moves, four diagonal moves, and a no-op:
$\mathcal{A}=\{\uparrow,\downarrow,\leftarrow,\rightarrow,\nwarrow,\nearrow,\swarrow,\searrow,\circ\}$.
Each agent observes $o_i \in \mathbb{R}^{6+(2r+1)^2}$ including normalized position, relative teammate position, relative goal position, and a local obstacle patch with radius $r=3$. We represent this compactly as:
$o_i=\big[\frac{p_i}{G},\frac{p_j-p_i}{G},\frac{g-p_i}{G},\mathcal{O}_i\big]$, where $p_i\in\mathbb{Z}^2$ is the agent position, $G$ is the grid size for normalization, and $\mathcal{O}_i$ is the flattened obstacle patch.

\subsubsection{Reward design}

The reward encourages cooperative goal-reaching while penalizing collisions and unsafe proximity to obstacles:
\[
R_i = R^{\text{goal}}_i + R^{\text{dist}}_i + R^{\text{step}}_i + R^{\text{col}}_i + R^{\text{obs}}_i + R^{\text{team}}_i
\]
In implementation, $R^{\text{goal}}_i$ provides a terminal bonus when $\|p_i-g\|<1.5$, $R^{\text{dist}}_i$ shapes progress toward $g$, $R^{\text{step}}_i$ discourages unnecessary motion, $R^{\text{col}}_i$ penalizes agent--agent collisions, $R^{\text{obs}}_i$ penalizes obstacle proximity when $d_{\text{obs}}<d_{\min}$, and $R^{\text{team}}_i$ rewards successful coordinated completion when all agents reach the goal. Episodes terminate when all agents reach the goal or a time horizon of 100 steps is exceeded.

\subsection{Trajectory Smoothing and Resampling}

To improve the interpretability and physical feasibility of trajectories produced by discrete grid navigation (important for real robot execution and human readability), we apply two post-processing steps: 

\subsubsection{Line-of-sight path smoothing}

Raw trajectories obtained from the environment often contain redundant waypoints, which are caused by discrete grid navigation and exploration noise. 
To address this, we adopt a line-of-sight-based smoothing method.
Given two waypoints $p_i$ and $p_j$, if the straight line segment connecting them does not intersect with any obstacle (checked via Bresenham's algorithm), then all intermediate waypoints between $p_i$ and $p_j$ can be removed. 
This reduces zig-zagging and produces shorter, more natural trajectories.

Formally, the smoothed trajectory is obtained as follows:
\[
\mathcal{P}_{\text{smooth}} = \{p_0, p_{k_1}, p_{k_2}, \dots, p_T\} 
\]
where each pair $(p_{k_m}, p_{k_{m+1}})$ satisfies the line of sight condition.


\subsubsection{Uniform resampling}

Since smoothed trajectories may contain irregularly spaced waypoints, we further perform a uniform resampling. 
Given a desired number of points $N$, the cumulative arc-length of the trajectory is first computed. 
Then, $N$ equally spaced distances are sampled along the curve, and the corresponding interpolated positions are generated:
\[
p^{(n)} = (1 - \lambda) \cdot p_{k} + \lambda \cdot p_{k+1}, \quad n=1,\dots,N
\]
where $\lambda$ is the normalized interpolation factor within the segment $[p_k, p_{k+1}]$.

This ensures consistent temporal alignment between agents, facilitating downstream coordination tasks such as collision avoidance and synchronized demonstrations.


\subsubsection{Visualization results}

\begin{figure}[!htbp]
    \centering \includegraphics[width=0.45\textwidth]{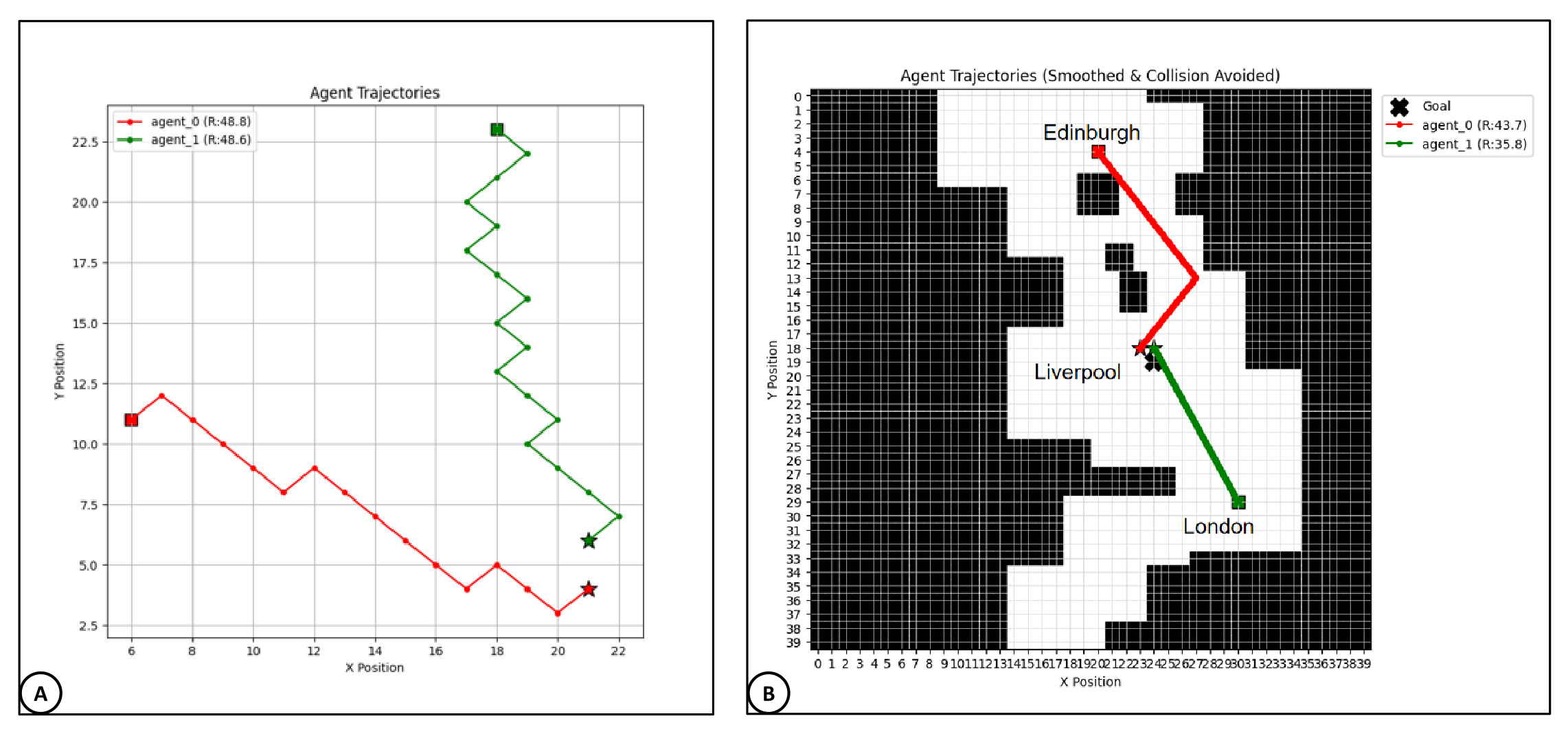}
    \caption{
    Trajectory post-processing for collision-aware and interpretable robot navigation: A) Raw MADDPG-generated trajectories in continuous x-y space, which may contain zig-zag motion caused by grid discretization and exploration noise, and B) Corresponding trajectories after line-of-sight smoothing and uniform resampling, overlaid on the UK grid-world map to produce safer, shorter, and more readable routes for city-to-city navigation.
}\label{fig:traj_compare}
\end{figure}

Figure~\ref{fig:traj_compare} illustrates the effect of our post-processing pipeline. 
The left panel displays the raw trajectories directly obtained from the environment, while the right panel illustrates the improved, smoothed, and uniformly resampled trajectories.



\section{EVALUATION and RESULTS} \label{Evaluation}



We evaluated one- and two-robot AcoustoBots on a scaled UK grid map. The evaluation targets (i) Physicalization fidelity (stable levitation and height variation during motion), (ii) Navigation performance (city-to-city traversal and coverage), and (iii) Safety and robustness (collision avoidance and stable behavior in shared space). We report both learning trends and real-world demonstrations, emphasizing behavior clarity and the interpretability of the particle-height cue.

\subsection{Model evaluation}

\begin{figure}[!htbp]
    \centering \includegraphics[width=0.43\textwidth]{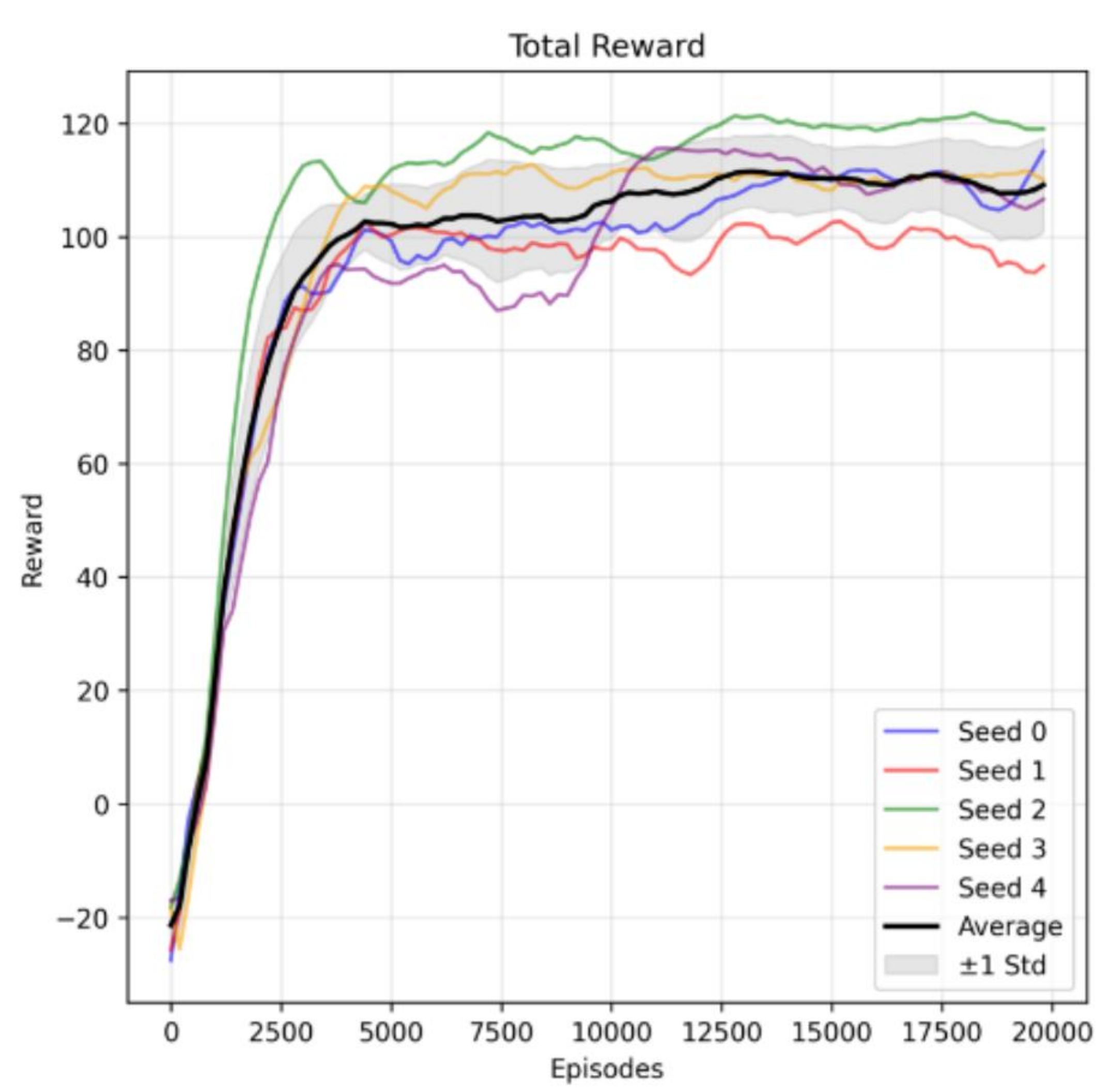}
    \caption{
    MADDPG training performance over 20,000 episodes across five random seeds. Colored curves show per-seed total reward, the black curve shows the mean reward, and the shaded region denotes ±1 standard deviation. Rewards increase rapidly during early training and then stabilize, indicating convergent and reproducible policy learning.
    }
    \label{fig:TotalReward}
\end{figure}

In this subsection, we analyze MADDPG training across five random seeds. The total reward captures the agents' ability to reach goals while avoiding penalties (e.g., collisions and inefficient motion). As shown in Figure~\ref{fig:TotalReward}, reward increases sharply within the first 2000 episodes, indicating that MADDPG quickly acquires basic goal-reaching and penalty-avoidance strategies. The curve then rises more gradually and converges to approximately 100--120 by 20{,}000 episodes, suggesting stable policy improvement. The relatively narrow variance across the five seeds indicates consistent learning behavior.
Additional metrics support the same trend: success rate increases and stabilizes around 80\% (with some seeds approaching 100\%), episode length decreases from $\sim$100 to $\sim$40--50 steps, and collisions converge close to zero. The mean distance-to-goal drops rapidly within the first 5000 episodes. Overall, training across five seeds yields an average total reward of $112.62 \pm 8.30$, indicating stable and reproducible learning.


\subsection{Experimental Setup}

\subsubsection{Environment}

\begin{figure}[!htbp]
  \centering
  \includegraphics[width=0.43\textwidth]{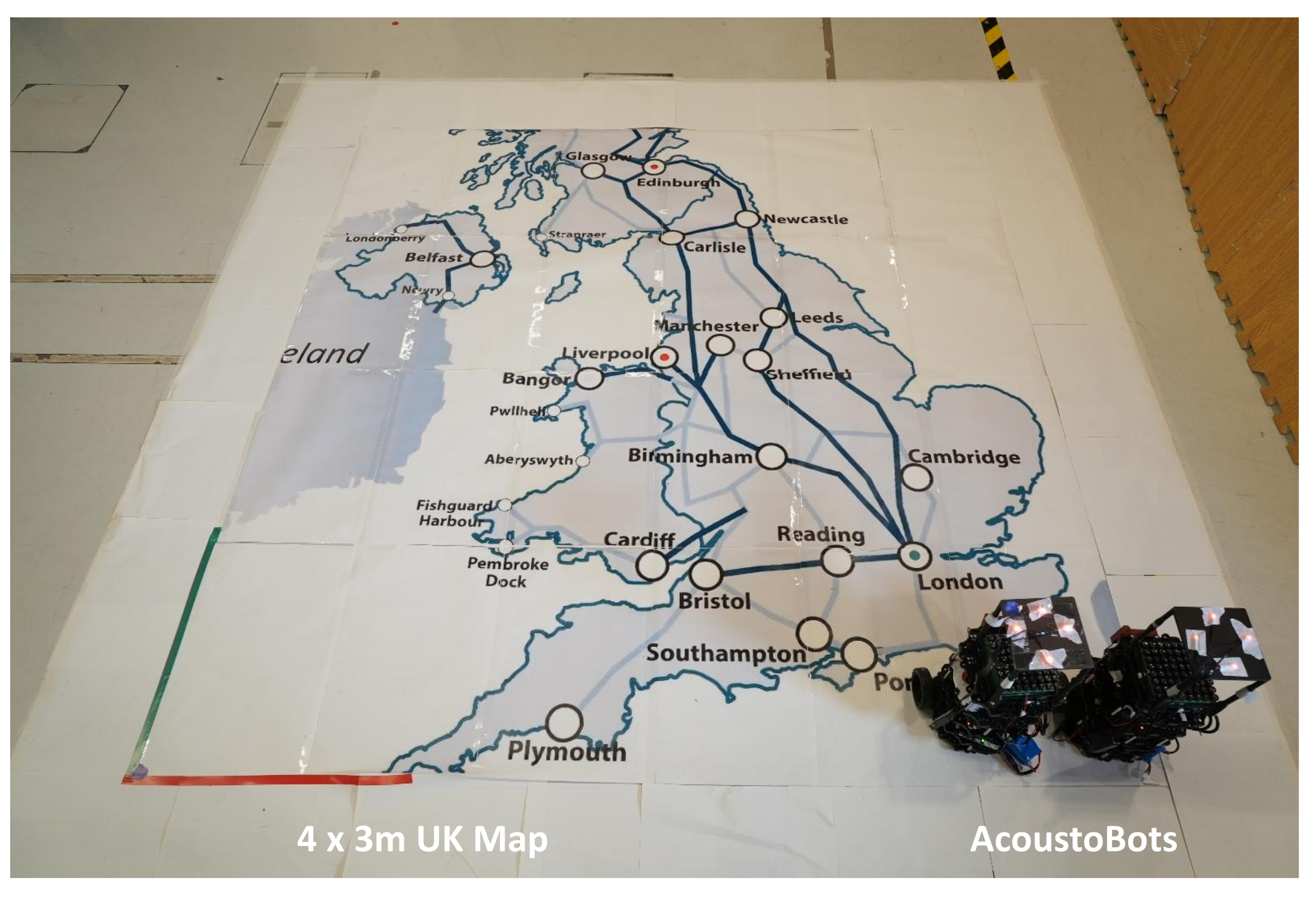}
  \caption{
  Experimental setup for the scaled UK navigation study. A 4 × 3 m UK map is deployed on the floor as a metric grid environment with labeled city waypoints. TurtleBot3-based AcoustoBots navigate between cities under the learned policy, enabling evaluation of city-to-city traversal and spatial data physicalization in the real-world testbed.
  }
  \label{fig:ExperimentalSetup}
\end{figure}

As shown in Figure~\ref{fig:ExperimentalSetup}, a map of the UK (physical size: $4 \times 3$ meters) is placed on the floor with metric grid cells of $\Delta = 10$ cm. City centroids $\mathcal{C}=\{c_k\}$ are placed in geodetically consistent locations. The scalar field $\phi$ (population) is sampled at $\mathcal{C}$ and bilinearly interpolated over the grid to provide a spatially continuous field for physicalization and navigation.

\subsubsection{Platform}

\begin{figure}[!htbp]
  \centering
  \includegraphics[width=0.43\textwidth]{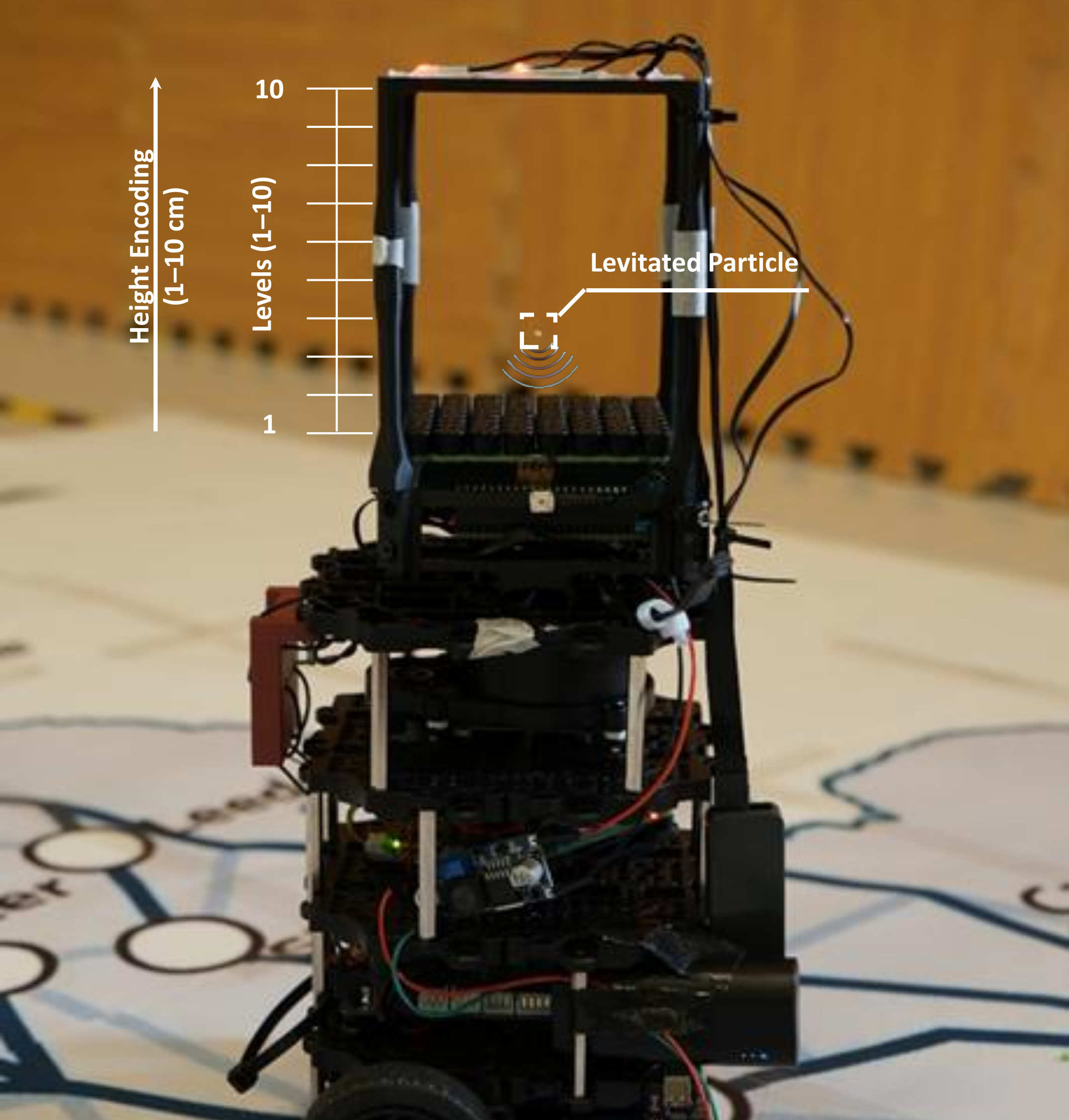}
  \caption{
  AcoustoBot demonstrating acoustophoretic data physicalization during navigation. While traversing the scaled UK map, the TurtleBot3-based platform maintains stable acoustic levitation, and the levitated particle height (1–10 cm) encodes the local scalar value at the robot’s current location. The annotated example shows a lower particle height, corresponding to a lower mapped value, illustrating in-motion physicalization of spatial data.
  }
  \label{fig:ExperimentalResult}
\end{figure}

Each AcoustoBot uses a TurtleBot3 base with wheel odometry and an upward-facing $8\times8$ PAT board (40 kHz), as shown in Figure~\ref{fig:ExperimentalResult}. During the experiments, the levitated particle provides the embodied physicalization cue, while the robot executes the learned navigation policy on the projected map.

\subsection{Experiments and Results}

We report results for two tasks: 
1) City-to-city traversal: 
London $\rightarrow$ Liverpool $\rightarrow$ Edinburgh (fixed waypoints) and
2) Team coverage: 
One or two robots are deployed across a set of target cities to maximize informative coverage under a specified path-length budget.

For Experiment \#1 (one AcoustoBot), we ran $N{=}10$ trials per regime, and for Experiment \#2 (two AcoustoBots) we ran $N{=}10$ trials per regime.

\subsection{Quantitative Summary}

Table~\ref{tab:summary} provides a compact overview of the key outcomes: success, safety (collisions), and demonstration-level interpretability (consistent execution of city-to-city and two-robot behaviors). 



\begin{table}[!htbp]
\centering
\caption{Summary of evaluation outcomes (mean values).}
\label{tab:summary}
\renewcommand{\arraystretch}{1.1}
\begin{tabular}{lcccc}
\toprule
Setting & Task & Success (\%) & Collisions & Trials \\
\midrule
1 Robot  & City-to-city & 90 & 1 & 10 \\
2 Robots & Coverage     & 80 & 2 & 10 \\
\bottomrule
\end{tabular}
\end{table}

\subsection{Discussion}

\begin{figure}[!htbp]
  \centering
  \includegraphics[width=0.43\textwidth]{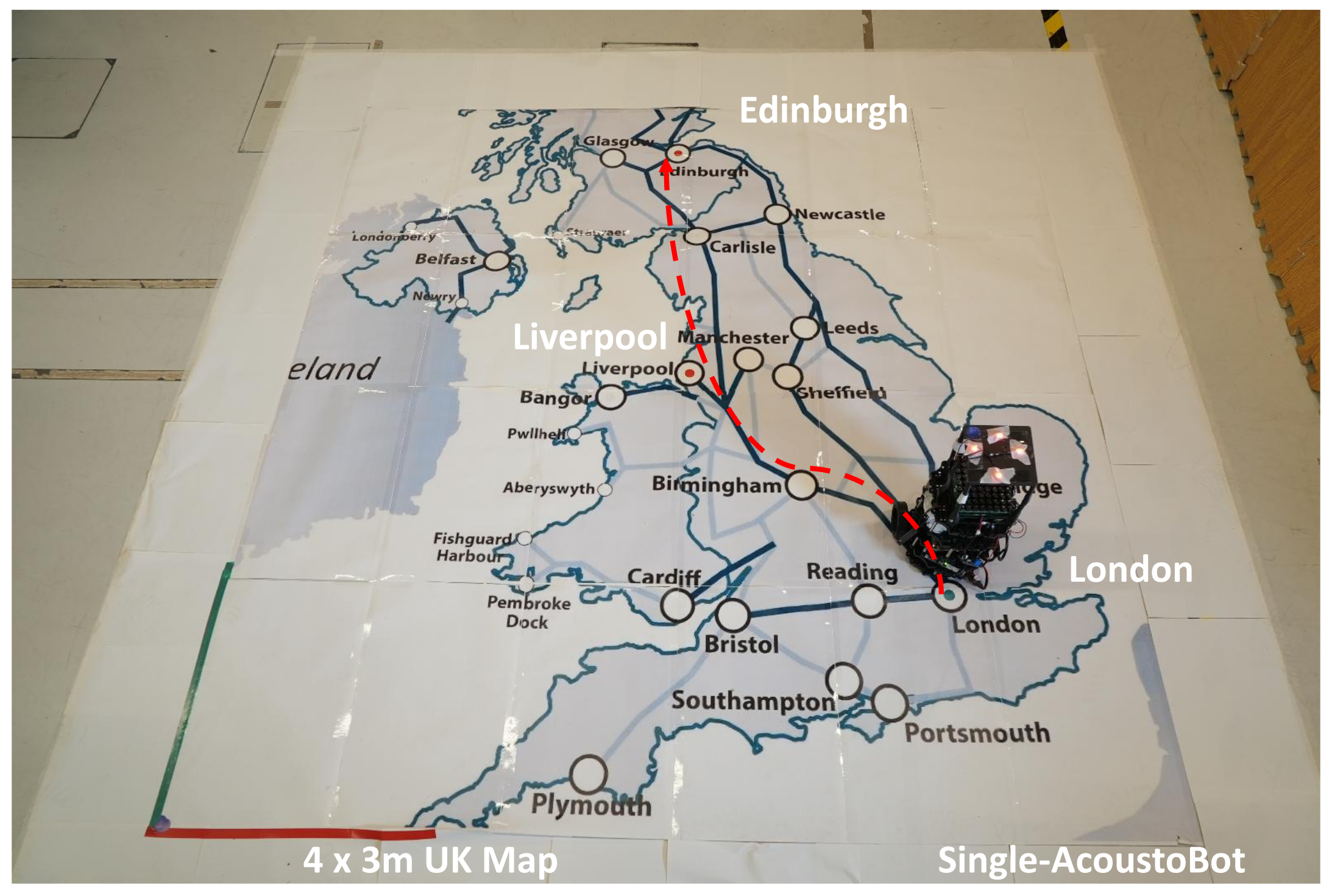}
  \caption{
  Single-AcoustoBot city-to-city traversal on the scaled UK map. The TurtleBot3-based platform navigates between labeled city waypoints under the learned policy, demonstrating geographically grounded and collision-aware motion in the real-world evaluation environment.
  }
  \Description{Single-AcoustoBot city-to-city traversal on the scaled UK map, showing the robot navigating between labeled city waypoints.}
  \label{fig:ExperimentalResult_One}
\end{figure}

\begin{figure}[!htbp]
  \centering
  \includegraphics[width=0.44\textwidth]{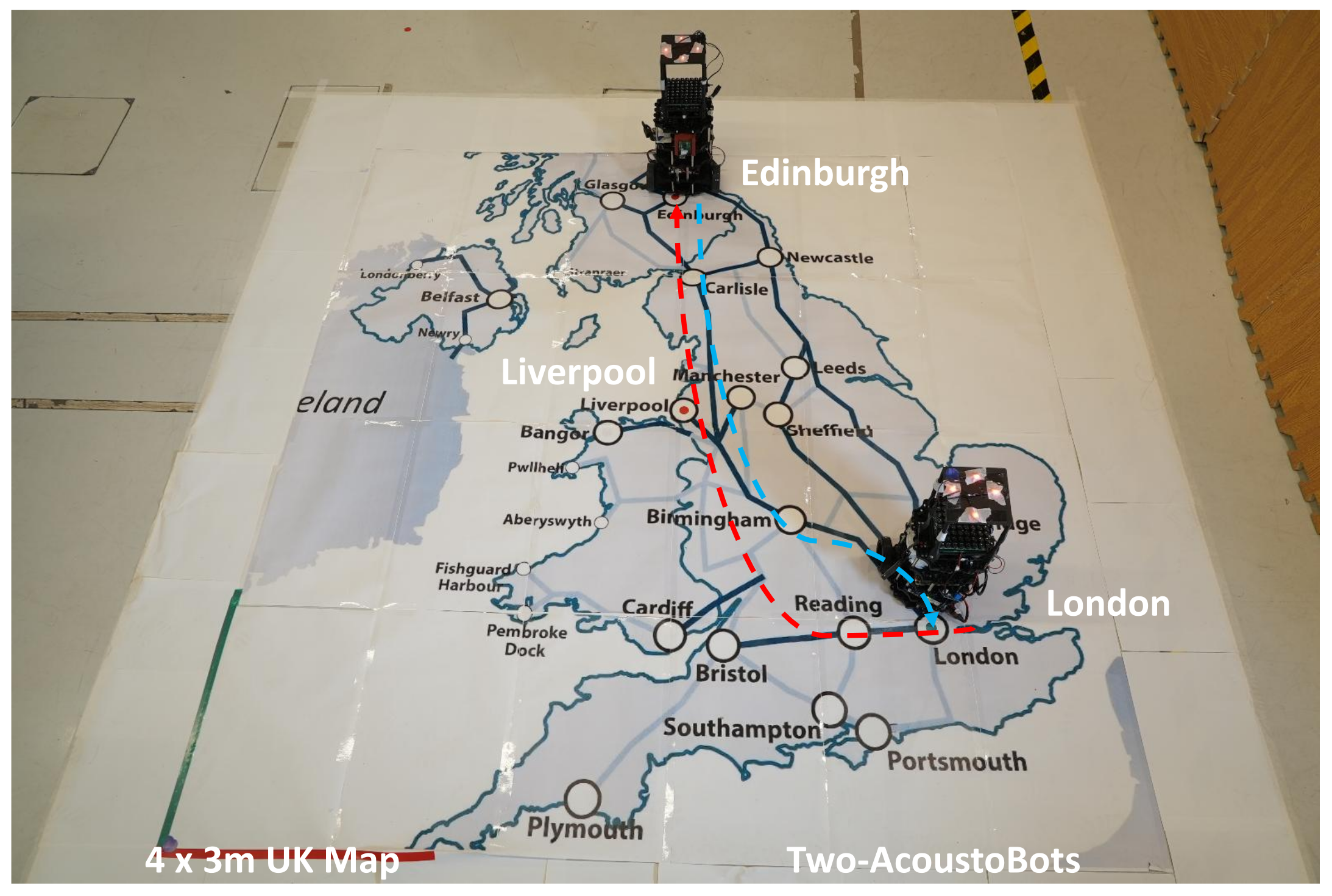}
  \caption{
  Two-AcoustoBot cooperative coverage on the scaled UK map. Two TurtleBot3-based platforms execute the learned multi-agent policy to coordinate motion across city regions while maintaining safe separation and avoiding collisions, demonstrating team-level coverage behavior in the real-world evaluation environment.
  }
  \label{fig:ExperimentalResult_Two}
\end{figure}

The results in Figure~\ref{fig:ExperimentalResult_One} and Figure~\ref{fig:ExperimentalResult_Two} indicate that the MARL policy produces stable, collision-aware navigation suitable for robot-mediated demonstrations, with trajectories that are smoother and easier to interpret as robots reach intended city waypoints.
Nevertheless, limitations remain. Success rates plateau around 80\% and 90\% rather than 100\%, and occasional mid-training fluctuations suggest sensitivity to exploration and environment stochasticity. Furthermore, experiments were conducted with a small number of robots in a simplified 2D grid-world abstraction, which may not capture the full complexity of larger teams or less structured real-world settings.
Future work will focus on scaling to larger and more heterogeneous teams, and improving reward shaping--potentially guided by human feedback \cite{miranda2023generalization}. In addition, integrating acoustic levitation with richer sensing (e.g., vision, audio, or environmental measurements) can enable layered, adaptive physicalization, positioning AcoustoBots as embodied agents for urban analytics \cite{chen2026cross}.

\section{CONCLUSIONS} \label{Conclusions}

This paper presented \emph{AcoustoBots}, a mobile acoustophoretic data physicalization platform that couples MARL with acoustic levitation to communicate spatial data through an embodied, glanceable cue. By mapping an urban scalar field (e.g., population density) to the height of a levitated particle (1--10\,cm) while robots navigate a scaled UK map, the system bridges sensing, decision-making, and communication in situ. From a human-robot interaction perspective, the key outcome is that the robot’s behavior and the physicalized signal remain co-located and interpretable, supporting robot-mediated demonstrations of geographically contextualized information.
Beyond this proof-of-concept, the approach suggests practical directions for robot-assisted decision support, for example in urban analytics where robots could traverse mapped environments and render local conditions (traffic flow, noise, air quality) as simple physical signals. Future work will focus on (i) scaling to larger teams with stronger robustness and safety guarantees (e.g., tighter collision avoidance and interference-aware coordination), (ii) improving levitation stability under motion and disturbances, and (iii) extending physicalization beyond a single height channel by integrating additional sensing modalities (vision/audio) and richer encodings. Together, these advances would strengthen AcoustoBots as adaptive, embodied agents for human-centered monitoring and communication in complex environments.






\section*{ACKNOWLEDGMENT}

This work was supported by the EPSRC Prosperity Partnership Program - Swarm Spatial Sound Modulators (EP/V037846/1), and by the Royal Academy of Engineering through their Chairs in Emerging Technology Program (CIET 17/18).





\bibliographystyle{IEEEtran}
\bibliography{biblio}

\end{document}